\documentclass[1p,preprint,12pt]{elsarticle}

\usepackage{amssymb}
\usepackage{amsthm}
\usepackage{amsmath}
\usepackage{algorithm}
\usepackage{algorithmic}
\usepackage{comment}
\usepackage{here}
\usepackage{hyperref}

\hypersetup{
  colorlinks=true,
citecolor=black,
linkcolor=black,
urlcolor=black
  }

\usepackage[dvipsnames]{xcolor}
\usepackage{todonotes}

\begin{document}




\title{Preservation of Feature Stability in Machine Learning under Data Uncertainty for Decision Support in Critical Domains}

\author[1,2]{Karol Capa{\l}a\corref{cor1}}
\ead{k.capala@sanoscience.org}

\author[1]{Paulina Tworek}
\ead{p.komorek@sanoscience.org}

\author[1]{Jose Sousa}
\ead{j.sousa@sanoscience.org}

\cortext[cor1]{Corresponding author}

\address[1]{Personal Health Data Science Group, Sano - Centre for Computational Personalised Medicine, Czarnowiejska 36, 30-054 Krak\'ow, Poland}
\address[2]{Institute of Computer Science, AGH University of Krakow, Kawiory 21, 30-059 Krak\'ow, Poland}

\begin{abstract}

In a world where Machine Learning (ML) is increasingly deployed to support decision-making in critical domains, providing decision-makers with explainable, stable, and relevant inputs becomes fundamental. 
Understanding how machine learning works under missing data and how this affects feature variability is paramount. 
This is even more relevant as machine learning approaches focus on standardising decision-making approaches that rely on an idealised set of features. 
However, decision-making in human activities often relies on incomplete data, even in critical domains. 
This paper addresses this gap by conducting a set of experiments using traditional machine learning methods that look for optimal decisions in comparison to a recently deployed machine learning method focused on a classification that is more descriptive and mimics human decision making, allowing for the natural integration of explainability. 
We found that the ML descriptive approach maintains higher classification accuracy while ensuring the stability of feature selection as data incompleteness increases. This suggests that descriptive classification methods can be helpful in uncertain decision-making scenarios.

\end{abstract}

\begin{keyword}

machine learning \sep critical scenarios \sep decision making \sep standardising \sep descriptive \sep  uncertainty \sep missing data
\end{keyword}



\maketitle

\section{Introduction}
\label{sec:intro}

Advancements in Machine Learning (ML) are changing decision-making in critical domains, from healthcare to finance~\cite{fu2021datadriven_decisionmaking}. 
This increases the need to provide models capable of being integrated into the decision-making process. 
However, despite their potential, ML algorithms are often plagued by limitations, including lack of generality and explainability, bias, and outcome stability under missing data~\cite{mitra2023missingdata}. 
Furthermore, ensuring stable and explainable predictions under missing data is crucial for establishing the reliability required for ML to support decision-making in critical domains. 
Traditionally, ML addresses missing data using imputation or modelling techniques~\cite{lin2020missingdatainputation}. 
These approaches come with two challenges. 
Its implementation requires data preprocessing, which can be costly and particularly time-consuming and challenging when using low-quality datasets. 
Secondly, imputing missing values can introduce erroneous information into the model, especially with high missing value rates or biased incomplete patterns.

Such challenges can cascade negative impacts on the entire model, weakening its ability to classify data correctly and potentially altering the identification of the most essential features. 

The adoption of ML methods to support decision-making in critical domains mainly focuses on increasing decision-making accuracy.
However, it needs to provide an explainable rationale to be integrated into the decision-making process. 
Therefore, an ML methodology that maintains feature stability under missing data~\cite{parr2017uncertainty}, while minimising the need for data preprocessing is crucial. 
Moreover, understanding the rationale behind a decision is fundamental for ML to support decision-making and forms the basis of explainable AI (XAI)~\cite{marcinkevics2020arxivexplainability,peters2023explainable,pronava2023explainability}.

To deal with the missing data without the need for imputation and data modelling, recently, an ML approach has been proposed by us~\cite{ibias2023sanda,gherardini2024cactus}. 
The proposed approach demonstrated the ability to maintain a constant level of classification accuracy even with a very high level of the data set incompleteness while being fully explainable and interpretable~\cite{gherardini2024cactus}. 

 Although the methodology consistently outperforms commonly used explainable methods with high levels of missing data, it performs noticeably worse with complete or almost complete data. 
 Therefore, it is important to verify if other abstraction methods affect the accuracy of the classification provided by the proposed ML approach and if, despite their change, the algorithm will remain highly resistant to missing data.

The variability of the most important features in the classification process is a significant problem for a model's explainability. 
Even if the missing data are randomly distributed, their presence in the dataset may considerably change the significance of individual features in the final classification. 
If the distribution of missing data is indeed random, this situation is highly undesirable, as one expects the model to reflect the actual population. 
Therefore, the ML sensitivity to changes that do not introduce bias into the data set, should be as low as possible.
To our knowledge, research on the stability of the features under data uncertainty in critical decision-making domains has been limited to date and has not been included in the previously proposed ML approach~\cite{gherardini2024cactus,ibias2023sanda}.

To sum up, this paper explores two new methods of abstraction to demonstrate the potential of data abstractions to preserve feature stability under missing data. 

The manuscript is structured as follows:
section~\ref{sec:relwork} details the ML algorithms employed in this research.
A discussion of the abstractions focusing on proposed alternatives to the previously proposed approach and their limitations is presented in section~\ref{sec:abstractions}.
The research problem and the experiments designed to investigate it are presented in section~\ref{sec:exp}, while section~\ref{sec:results} reports the obtained results and their subsequent discussion.
The manuscript concludes with a summary and conclusions in the section~\ref{sec:summary}.

\section{Related work}\label{sec:relwork}
The aim of the following study is to evaluate machine learning methods for critical decision-making when available datasets contain a large amount of missing data. 
Therefore, it is crucial to select methods for comparison that can perform effective classification while also providing information about the most significant features. 
This section, thus, presents two commonly used methods in this context, along with our previously proposed approach.

A similar question has been posed in the context of feature selection~\cite{liu2020feature}. However, these studies focused solely on selecting the most important features to reduce dimensionality for model construction, rather than emphasizing explainability using existing models. Additionally, the data was supplemented using imputation methods.

\paragraph{Random Forest (RF)\label{sec:RF}} RF was originally presented as an approach for improving the accuracy of a single decision tree. However, sometimes it has problems with over-fitting and therefore with generalisation~\cite{belle2021principles}.
RF~\cite{breiman2001random} is based on an ensemble of individual decision trees trained on random samples of the training data, thereby achieving different characteristics of the data distribution.
Once the decision trees have been constructed, the RF algorithm makes a prediction by averaging the predictions of all the trees, reducing the model's variance.
This often produces more accurate predictions than any individual tree could make while remaining explainable.
Moreover, in the case of medium-sized, tabular datasets, RF may outperform deep learning on classification task~\cite{gill2022machine,grinsztajn2022tree}.
These two properties make it suitable for use in various sensitive areas.

\paragraph{Boosting Trees (BT)\label{sec:Boost}} BT is an alternative approach to tree-based algorithms. It is included among other boosting techniques~\cite{friedman2001greedy}. 
Similarly to RF, it is intended to enhance the performance of a decision tree. 
One of the most popular types of boosting algorithms is Gradient Boosting (GB).
GB is an ensemble learning technique that builds a sequence of decision trees, each trained to correct the errors of the previous one while minimizing the objective function.
This method combines multiple weak decision trees to form a stronger overall model.
Other examples of boosting algorithms are XGBoost~\cite{chen2016xgboost} and Light Gradient Boosting Machine (LGBM)~\cite{ke2017lightgbm}.
In this study, GB was chosen because it scales better than both XGBoost and LGBM. 
Furthermore, its prioritized simplicity and interpretability make GB potentially more suitable for the proposed experiments.

Both RF and BT are ensemble methods characterised by lower explainability than single decision trees. However, they can still provide features important for classification, which makes it possible to clarify on what basis the decision was taken.

\paragraph{Previously proposed approach (PPA)\label{sec:sanda}}%
The PPA consists of two main parts, which interact with each other to provide a comprehensive method of data analysis: classification and knowledge graphs (KG)~\cite{ibias2023sanda,gherardini2024cactus}.
While the classification module can separate data into classes and provide information about how well the model captures the data's properties, KG enhances the explainability of the results and provides deeper insight into the interdependence between features. 
Since the primary objective of this paper is to investigate the potential for enhancing previously proposed data representation capability, which is primarily evaluated through the classification aspect, this section focuses on explaining the employed classification method. 
The key concepts utilized for the classification problem include abstractions and the classification algorithm itself.

The PPA employs a discretisation protocol called abstractions, which uses receiver operating characteristic curve (ROC) theory for binarization of the data~\cite{ibias2023sanda,gherardini2024cactus}.
This transformation is applied to data with continuous distributions or discrete data that assume more values than desired.
Such abstracted data can be combined with the natively categorical data, enabling easy integration of different types of data~\cite{gherardini2024cactus}.
A more detailed description of ROC abstraction is provided in Sec.~\ref{sec:abstractions}.

Based on obtained data abstractions, an explainable KG representation~\cite{jin2020selfsupervised, ibias2023sanda, gherardini2024cactus}. 
Through building the graphs, a representation for each class based on available features is prepared. 
The significance of each vertex of the KG is represented by the probability of each feature being in a given category of the given class. 

The work focused on binary classification, however the ML algorithm can be described and performed also for arbitrary number of classes~\cite{ibias2023sanda,gherardini2024cactus}.
Let $X=\{x_1,\cdots, x_n\}$ be the set of abstracted features and $C_j$ be the $j$th class. 
Then for given $C_j$ the following probability can be computed
\begin{equation}
    P(X|C_j)= \prod_{i=1}^n P(x_i|C_j)
    \label{eq:probability}
\end{equation}
The class with the highest $P(X|C_j)$ is chosen, i.e.
\begin{equation}
  \mathrm{argmax}_C\left\{ P(X|C_j) \right\}.
  \label{eq:max_probability}
\end{equation}
Therefore, it assigns a given set of values to the class for which its occurrence is most likely.
If the given feature $x_i$ is empty (missing value) or is represented by null values, it is skipped in the calculation of the probability $P(X|C_j)$.

Contrary to most commonly used methods, the PPA does not divide feature space based on the optimisation of some classification functions. 
Instead, the values of every feature are individually divided based on abstraction methods, which do not need to, but can, take into account class distribution. 
From the perspective of the entire feature space, it creates division into a grind, in which every ``cell'' is labelled as one of the classes based on the conditions given by Eqs.~\eqref{eq:probability} and \eqref{eq:max_probability}.
These conditions do not consider every individual ``cell'' but rather try to approximate it, base on every dimension separately. 
On one hand, this approach may diminish the method's classification accuracy by failing to incorporate more intricate, nonlinear relationships, particularly in the context of a sparse grind, where the number of abstractions is low.
On the other hand, it allows the classification test to be completed even in the presence of missing values by considering lower-dimensional space.
Let $x_{k,f}$ denotes $f$th value of the $k$th feature and $C_i$ is as previously $i$th of $N$ classes. 
The importance of the $x_{k,f}$ on the classification result can be determined through the measure, called feature rank~\cite{gherardini2024cactus}
\begin{equation}
    R_{x_{k,f}} = \frac{2\sum_{i=1}^{N} \sum_{j>i}^{N} |P(x_{k,f}|C_i) - P(x_{k,f}|C_j)|}{N(N-1)},
    \label{eq:rank}
\end{equation}
which describes how much the probability of the given value differs between classes and, therefore, how much this value influences classification.
The importance of the feature can be determined by averaging rank over all values of the feature
\begin{equation}
    \bar{R}_{x_k} = \frac{1}{|x_k|}\sum_{f \in x_k} R_{x_{k,f}}.
    \label{eq:avg_rank}
\end{equation}

\section{Abstractions}
\label{sec:abstractions}
Abstractions constitute a fundamental element of PPA, hence it is important to discuss and analyze them comprehensively. 
This section is dedicated to presenting various abstraction methods and discussing their limitations.

The abstraction process maps original values from the data into a smaller set of values.
To this end, data abstraction creates a simplified representation of the underlying data, while hiding its complexities and associated operations.
In addition, the abstraction method used ensures the anonymisation of the data. 
It is important to mention that the abstraction process is performed on a column-by-column basis. 

In the previous works, only binary discretisation based on the ROC curves was considered as a part of the ML algorithm~\cite{ibias2023sanda,gherardini2024cactus}.
Such a choice of abstraction is both limited in method division criteria as well as the number of categories into which data is transformed. 
Both of these limitations can significantly influence the decision-making process, as abstractions form the basis for how the algorithm divides space.
This paper explores two alternative methods of data abstraction. 
These methods focus on different properties of data distribution and can create any number of abstraction categories.

\subsection{ROC Curves}
ROC curve abstractions~\cite{ibias2023sanda} split the values of a feature into two categories in a way that maximises the separation between the classes in the feature. 
To be more precise, let $F_1(x)$ be the cumulative density function (CDF) of the $1$st class and $F_2(x)$ CDF of the $2$nd class.
Difference between populations of both classes below $x$ is given by
\begin{equation}
\Phi(x) = |F_1(x) - F_2(x)|.
    \label{eq:deltaCDF}
\end{equation}
ROC curve abstractions aim to maximise this difference, i.e., finds $x$ which is a global maximum of $\Phi(x)$.
It can be generalised for multi-class classification by one versus all division, or by choice of one of the pair-wise created boundaries.

\subsection{static binning}
Abstractions, through static binning, divide the values of a feature into categories of equal range. 
Let $A$ be a set of the values taken by the feature. 
Then $L=\mathrm{max}A - \mathrm{min}A$ is a range of values. 
static binning divides the range of the data $L$ into $n$ equal intervals of length $L/n$, assigning the feature a number corresponding to the bin number in which its original value falls.
Edge bins can be used to incorporate values that fall outside the range present in the training dataset.

Constant bins are independent of the class distribution for a given feature, thus they can be applied to both binary and multi-class tasks without additional modifications.

\subsection{Quantiles}
While static binning divides data to keep the range of every bin equal, abstractions based on quantiles divide data into categories of equal size. 
Cut-off values are chosen based on the values of quantiles.
The quantile of the order $n$ is defined based on the probability distribution of the random variable ranging over the set $X$ in such a way that the locations $y_q$ of the quantile $q$ ($q=k/n$, where $k\leq n$) given by 
\begin{equation}
 q = \int_{-\infty}^{y_{q}} p(x) dx.
 \label{eq:quantile}
\end{equation}
In other words, quantile $y_q$ is such that the cumulative distribution function takes at this point value $q$, or alternatively, the $q$th part of the data takes values smaller or equal to $y_q$.
Quantile-based abstractions, therefore, assign the feature the largest $k$ for which $x<y_q$, i.e., corresponding to the highest quantile larger than the initial value.  
The most commonly used quantiles are ones of the order $n=2$ (median), $n=4$ (quartiles) and $n=10$ (deciles).

Similarly to static binning abstractions, quantiles are also class-agnostic, allowing them to be employed for multi-class problems without modifications.

\subsection{Limitation on number of abstractions}
\label{sec:abs_limit}
In this work, we propose an approach that takes into account the partitioning of the data of a given characteristic by means of abstraction into 10 or 20 different abstractions. 
However, quantile-based abstraction methods can theoretically generate abstractions down to the size of the original data, while static binning is not even limited by this.
It should be taken into consideration, that an increase in the number of abstractions reduces the statistical significance of their occurrence probabilities.
In other words, when there are too many abstractions compared to the total amount of data, the abstractions are unable to accurately reflect reality.
Therefore, it is important to determine the extent to which the obtained probability deviates from a purely random result, and the observed differences in probabilities are inherent to the dataset itself rather than to the analysed phenomenon.

Let $c$ be the number of classes, $N$ population size and $n$ number of bins created by abstractions. 
The expected probability of each class in a given bin is a uniform probability $p=1/c$ for the purely random case.
However, to properly select the number of abstractions, it is more important to estimate the accuracy with which we can distinguish a random result from a true one, i.e., the standard deviation.
Since our estimate of probability is the ratio of the selected class to the total population, the standard deviation is
\begin{equation}
    \sigma=\sqrt{\frac{p(1-p)}{\mathcal{N}}},
\end{equation}
where $\mathcal{N}$ is total population (number of items) in the given bin. 
For quantile-based abstractions, the population in each bin is constant, i.e., $\mathcal{N}=N/n$.
Substituting the symbols into the formula we get:
\begin{equation}
    \sigma=\sqrt{\frac{\frac{1}{c}(1-\frac{1}{c})}{\frac{N}{n}}}=\sqrt{\frac{(c-1)n}{c^2N}}.
\end{equation}
It is desirable to minimise $\sigma$, but this would mean minimising the number of abstractions, which may ultimately affect the model's performance.
Therefore, it is necessary to assume the separation with which we can distinguish a random result from a significant one. 
Therefore, for the adopted resolution R, equality must follow
\begin{equation}
    R=z\sigma,
    \label{eq:res_eq}
\end{equation}
where $z$ is a constant related to how certain the result should be.
After solving Eq.~\eqref{eq:res_eq} for the number of bins, one can find the upper limit for the number of abstractions
\begin{equation}
n_{\mathrm{max}}=\frac{c^2 R^2}{(c-1) z^2} N .
    \label{eq:max_no_bins}
\end{equation}
For binary classification Eq.~\eqref{eq:max_no_bins} simplifies to
\begin{equation}
n_{\mathrm{max}}=\frac{4 R^2}{z^2} N.
\end{equation}
For constant bins, the equation can also be used, remembering that it is based on the average of the bin population.

\section{Experiments}\label{sec:exp}
As highlighted in Section~\ref{sec:intro}, PPA~\cite{ibias2023sanda,gherardini2024cactus} is a promising candidate due to its stability with increasing data incompleteness.
Therefore, this research initially explores methods to improve the accuracy of PPA predictions by changing the abstraction method.
Subsequently, PPA is compared with state-of-the-art explainable ML methods regarding classification performance metrics and stability of feature importance.

The measures used are presented in section~\ref{sec:ver_methods} and the datasets in section~\ref{sec:datasets}. 
A detailed description of the experiments is provided in section~\ref{sec:exp_plan}.

\subsection{Verification metrics}
\label{sec:ver_methods}
Several metrics are used as a valuable tool for comparing the performance of ML classification models. 
They provide a comprehensive assessment of a model's performance across different aspects.
To define chosen metrics the following basic concepts describing classification results should be introduced:
\begin{itemize}
\item True positives ($\mathrm{TP}$) - number of correctly classified instances of the positive class,
\item False positives ($\mathrm{FP}$) - number of incorrectly classified instances of the negative class,
\item True negatives ($\mathrm{TN}$) - number of correctly classified instances of the negative class,
\item False negatives ($\mathrm{FN}$) - number of incorrectly classified instances of the positive class.
\end{itemize}
Based on the concepts several metrics of classification tasks are defined.
Among these, balanced accuracy (BA) provides a comprehensive assessment of model performance. 
BA is the primary metric used to evaluate the performance of selected abstraction methods and compare them with RF and GB. 
It is one of the simplest metrics for evaluating a classification model's ability to accurately predict classes in the context of imbalanced datasets, which are common, among others, for medical and financial problems.
BA is an extension of standard accuracy metrics, it is an average accuracy from both the minority and majority classes, i.e.
\begin{equation}
\mathrm{BA} = \frac{\frac{\mathrm{TP}}{\mathrm{TP}+\mathrm{FN}}+\frac{\mathrm{TN}}{\mathrm{TN}+\mathrm{FP}}}{2}.
\label{eq:BA}
\end{equation}

Beyond a simple BA, there are other metrics which offer different insights into how well a classification model is performing~\cite{hicks2022evaluation}.

Recall, defined as
\begin{equation}
\mathrm{recall} =\frac{\mathrm{TP}}{\mathrm{TP} + \mathrm{FN}},
\label{eq:recall}
\end{equation}
measures the ability for the correct identification of instances of the positive class from all the actual positive samples in the dataset. 
This metric is widely employed in various domains, particularly in application classification algorithms used, e.g. in medical and financial scenarios. 
For instance, in healthcare, it is crucial, especially in medical screening and diagnostic testing, as the high value of recall suggests that the test performs well in detecting $\mathrm{TP}$ cases, thereby reducing the chances of missing instances ($\mathrm{FN}$)~\cite{hicks2022evaluation}. 
In the financial area, recall is frequently measured in the context of risk management and fraud detection to ensure that potential risks or fraudulent activities are detected effectively~\cite{naoufal2020fraud}.

However, there are also scenarios where the accuracy of positive predictions made by a classification model is critical. For example, the consequences of carrying out medical or financial interventions or procedures that are not actually required, stemming from false positive results, can have a substantial impact. 
In such cases, precision should be monitored. 

Precision provides information about the quality of the model's positive predictions. 
It is defined as the number of positive instances divided by the total number of positive predictions, both correctly and incorrectly classified as positive class:
\begin{equation}
\mathrm{precision} =\frac{\mathrm{TP}}{\mathrm{TP} + \mathrm{FP}}
\label{eq:precision}
\end{equation}

Achieving a equilibrium between recall and precision holds great importance in many applications. 
However, the optimal metrics for the evaluation of ML models should be chosen based on the specific scope and nature of the problem at hand~\cite{powers2007evaluation}.

The influence of missing data on feature importance can be effectively assessed by directly comparing feature importance scores. 
However, this approach hinders comparisons across models and becomes cumbersome for analysing numerous datasets or datasets with high-dimensional features.

To facilitate a comparative analysis of features' significance stability, a quantitative measure is required. 
Therefore, we introduce the concept of relative feature significance change.
Let $\Sigma_{i,k}(d)$ be importance of the $i$th feature in $k$th model for percentage of missing data $d$.
Relative feature significance change can be defined as
\begin{equation}
    \Xi_{i,k}(d) = \frac{|\Sigma_{i,k}(d) - \Sigma_{i,k}(0) |}{\langle \Sigma_{i,k}(d) \rangle_d}.
    \label{eq:relative_significance}
\end{equation}
The form of the numerator is dictated, on one hand, by the fact that the quantity we are interested in is the change relative to the model without missing data and, on the other hand, by symmetry due to the sign of the difference.
$\langle \Sigma_{i,k}(d) \rangle_d$ denotes the average over different levels of missing data. 
This form of the denominator is intended to normalise the difference to the expected order of magnitude of the feature's importance. 
Its purpose is to ensure that a small percentage change in an important feature does not overshadow significant changes in less important features.
An alternative natural choice would be to use significance with full data for normalisation $\Sigma_{i,k}(0)$.
However, this choice breaks the symmetry associated with the sign of the difference, strengthening cases when $\Sigma_{i,k}(0) < \Sigma_{i,k}(d\neq 0)$.
Another approach may use normalisation to the highest obtained feature importance, but this has been rejected due to its potentially greater susceptibility to outliers.
Additionally, to facilitate comparisons of model robustness to missing data, the relative feature significance change metric, $\Xi_{i,k}(d)$, can be averaged over the number of missing data and features as well as its standard deviation.

\subsection{Datasets}
\label{sec:datasets}
For the sake of comparison with the original research, we used the $5$ datasets from SaNDA~\cite{ibias2023sanda} in the following experiments, supplemented by $10$ synthetic DIGEN~\cite{orzechowski2022generative} datasets (8\_4426, 10\_8322, 17\_6949, 22\_2433, 23\_5191, 24\_2433, 32\_5191, 35\_4426, 36\_466, 39\_5578). 
A selection of datasets is briefly described below.

The Ionosphere dataset~\cite{data, Sigillito1989Ionosphere}, a collection of $351$ radar measures of the ionosphere in Goose Bay, Labrador, is used to classify its structure. 
The dataset includes $34$ numeric features that measure the number of free electrons and other electromagnetic signals in the ionosphere.

The Sonar dataset~\cite{data, sonar} consists of $60$ numerical features that measure the shape and characteristics of the sonar signal. 
The classification task is to distinguish underwater surfaces as rock or metal. 
It is the smallest number of records equal to $208$.

The Wisconsin Breast Cancer dataset~\cite{data, Wolberg1995Breast} uses $30$ numerical features that measure the shape and composition of a breast mass to describe $569$ fine needle aspirates. 
The task is to distinguish between cancerous and non-cancerous samples. 

The Accelerometer dataset~\cite{data, Scalabrini2019Prediction} was generated for the prediction of motor failure time with the application of an artificial neural network. 
It uses $4$ numerical features; $3$ of them represents the values of x, y and z axes, while the fourth is cooler maximum speed percentage ranging from $20$\% to $100$\% with $5$\% intervals. 
The fifth attribute was used as the target class, where normal configuration was admitted as a negative class, while perpendicular and opposite configuration was a positive class. 
It has the largest number of records equal to $102000$.

The HIGGS dataset~\cite{data, baldi2014searching} is produced from Monte Carlo simulations of particle decays and contains $1.1\times 10^7$ entrances. 
Each process is described by $28$ features, of which $21$ are kinematic properties measured by the particle detectors in the accelerator and the remaining $7$ are quantities derived from them. 
The classification task is to distinguish between measurements of background noise and those connected to the observation of the Higgs particle. 
$10^5$ randomly selected records were taken for the experiments presented in this work.

DIGEN~\cite{orzechowski2022generative} datasets were designed to differentiate the performance of some of the leading classification methods. It is the collection of $40$ synthetic datasets created from each of the generative mathematical functions for testing binary classification tasks. 
Every dataset from DIGEN contains $10$ features of $1000$ normally distributed values.

\subsection{Experiment design}
\label{sec:exp_plan}
\subsubsection{Choice of abstractions}
\label{sec:exp_abstr}
The first experiment focuses on exploring the impact of the different abstraction methods presented in Sec~\ref{sec:abstractions} on PPA performance with an indication of classification accuracy.
For this purpose, each of the studied datasets was transformed using one of the following abstraction methods:
\begin{itemize}
    \item static binning into 10 or 20 bins
    \item quantiles of order 10 (deciles) or 20
    \item using the ROC curve method as a control group
\end{itemize}
Then, using the classification algorithms from PPA, models for selected datasets transformed by presented abstraction methods were created, and classification was evaluated.
It is important to note that in this experiment we do not create any missing data in the datasets. 

\subsubsection{Classification performance}
\label{sec:classifiaction_exp}
The second experiment focuses on the comparison between commonly used explainable ML methods -- RF and GB with PPA, which in previous studies claimed high stability of balanced accuracy value in respect to percentage of missing data~\cite{ibias2023sanda,gherardini2024cactus}. 
Contrary to the aforementioned studies, this experiment attempts to analyse not only the BA of the tested models but also recall and precision, which are dictated by their importance for the critical decision-making (CDM). 
Moreover, the PPA algorithm was modified through the different abstraction protocols, selected based on the results of the experiment described in Sec.~\ref{sec:exp_abstr}, and GB was included as less explainable but able to identify the most relevant features for classification. 
Such a choice of tested models and metrics allows for a more comprehensive examination of methods capable of supporting decision-making in critical domains. 

To examine their performance over different levels of data incompleteness, RF, GB and PPA were applied to datasets containing $1\%$, $5\%$, $10\%$,  $20\%$ and $50\%$ of missing data. 
Datasets with $n\%$ missing data were created from the original datasets by randomly removing $n\%$ of data from each column (feature). 

\subsubsection{Feature significance}
As feature significance is directly linked to the explainability of the classification process~\cite{pronava2023explainability}, this experiment is an extension of one presented in the section~\ref{sec:classifiaction_exp}.
For each selected model, both with and without missing data, the most important features for the classification process were extracted.
In the case of PPA, these were ranks defined by Eq.~\eqref{eq:avg_rank}.
For RF and GB, the importance of features was calculated using a built-in method in sci-kit learn after training the classification model.
It is referred to as the impurity-based feature importance, and it represents how strongly the feature helps to reduce impurity and improve classification accuracy. 
The feature's importance is determined as the normalised total reduction in the criterion that the feature contributes. This measure is also referred to as Gini importance defined as the average increase in purity from splits involving a given variable~\cite{scikitlearnfeaturesimportance}. If a variable is useful, it tends to split mixed-label nodes into pure, single-class nodes. Splitting based on permuted variables typically does not affect node purity. 
In a single growing decision tree included in RF or GB it can be presented as the dataset is repeatedly split based on the features to create a tree-like structure. 
Impurity is measured at each split node using the Gini metrics. 
Once the tree is constructed, feature importance is determined by how often each feature was used to split nodes and the reduction in impurity it achieved ~\cite{bjoern2009comparison}.

\section{Results}\label{sec:results}
\subsection{PPA abstraction method update}

\begin{figure}[!h]
    \centering
    \includegraphics[width=0.85\linewidth]{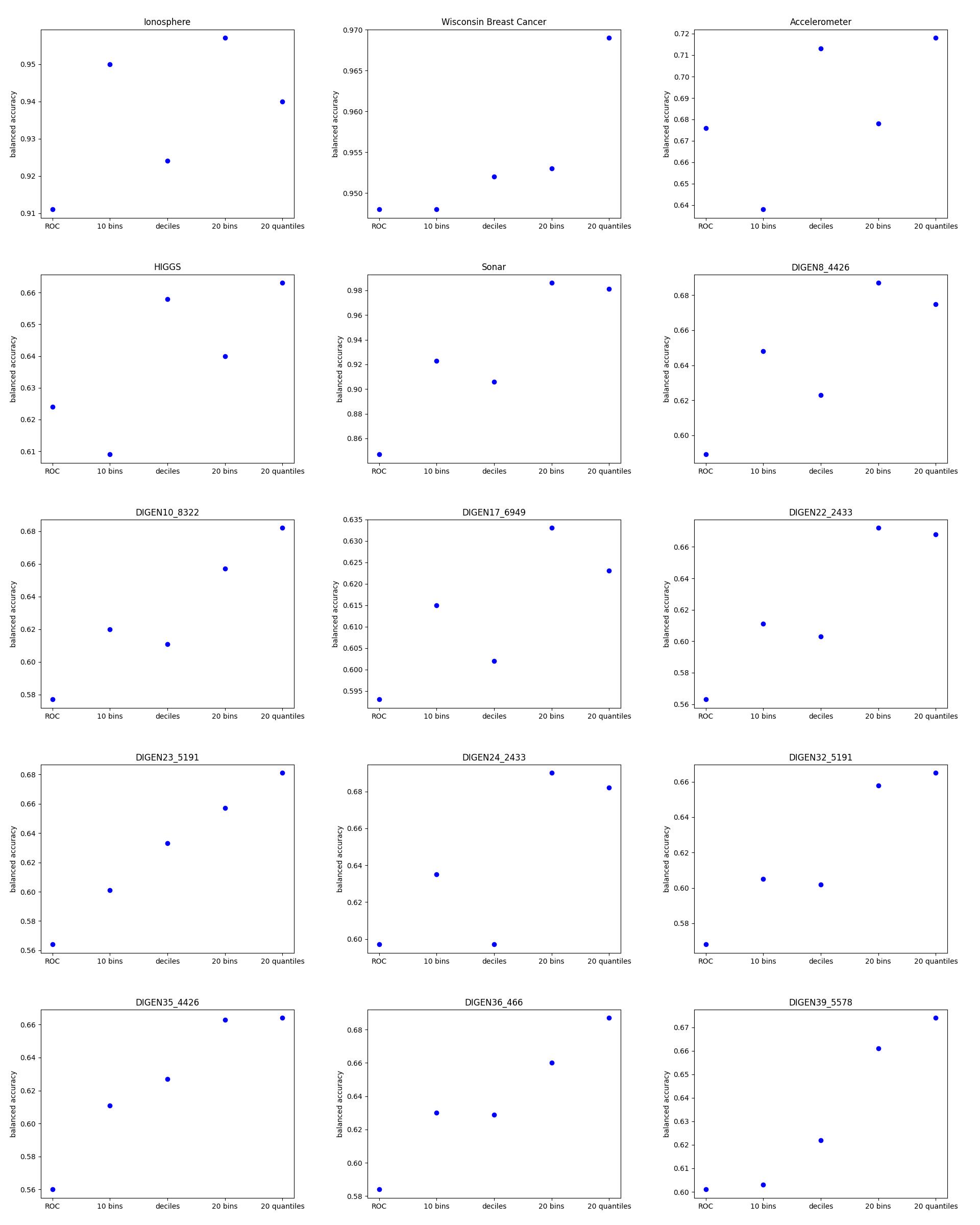}
    \caption{Balanced accuracy (BA) of the PPA classification method as a function of the different abstraction methods.}
    \label{fig:BA_abstractions}
\end{figure}

Figure~\ref{fig:BA_abstractions} shows the BA for experiment 1. 
For each dataset detailed in Section~\ref{sec:exp}, it contrasts the BA acquired from SaNDA across various abstraction types: ROC, deciles (quantiles $10$), quantiles $20$, $10$ bins, and $20$ bins.
For the WDBC, accelerometer, Higgs and DIGEN datasets, with the exception of DIGENs: 22\_2433, 24\_2433, and 17\_6949, the use of quantiles $20$ results in superior BA for classification.
On the other hand, for the remaining DIGENs, Sonar and Ionosphere datasets, the application of $20$ bins abstraction yields higher BA compared to other abstraction methods.

Based on this experiment, it can be observed that overall quantiles $20$ offers better classification performance compared to $20$ bins abstraction method. 
In the case of the opposite being true the discrepancy is significantly narrower.
A similar, but weaker trend can also be observed between deciles and $10$ bins.
In light of the results, quantile-based abstractions are a preferable abstraction choice for PPA classification.

However, the number of categories into which data is transformed as a result of the abstraction process is of far greater significance than the type of abstraction employed.
This can be explained by the fact that a higher number of abstractions (categories) enhances the resolution of the grid created in the feature space. 
As a result, a denser grid enables a more homogeneous distribution of data across its constituent parts, leading to more accurate classification performance.

One might naively believe that increasing the number of abstractions will enhance classification accuracy.
However, this approach harbours potential pitfalls that warrant careful consideration. 
First, increasing the number of abstractions can result in a decrease in the average number of data points per category, potentially compromising the accurate representation of statistical properties, as discussed in Sec.~\ref{sec:abs_limit}.
Furthermore, an increase in the number of abstractions can lead to an expansion in the number of nodes within the KG generated by PPA, potentially diminishing its explainability.
Finally, an extensive number of abstractions can exert a detrimental impact on the computational time required for generating classification and KG, a factor that may become particularly crucial when computational resources are constrained or the analysed problem demands expediency.

Consequently, the selection of the number of abstractions represents a trade-off between achieving high accuracy and managing uncertainties.
To determine the optimal number of abstractions, one should consider the specific requirements of the problem at hand, particularly the acceptable level of error tolerance and desired level of accuracy. 
Based on these considerations, values for the parameters $R$ and $z$ can be established, and the number of abstractions can then be calculated using Eq.~\eqref{eq:max_no_bins}.
Since the choice of parameters is individual to each project's needs, a constant number of abstractions was adopted in accordance with the one analysed in this section for easy comparison in the rest of the article.
At the same time, adopting a larger number of abstractions for smaller datasets would lead to an undesirable reduction in the certainty of the obtained probabilities.
It should be noted, however, that exceeding the number of abstractions determined from Eq.~\eqref{eq:max_no_bins} does not automatically mean that the obtained results will be incorrect.
However, in this case, additional tests are needed to verify the generalizability of the model.

\subsection{Classification Performance in the Presence of Missing Data}
Having identified the most effective abstraction method for PPA, the next step is to evaluate its ability to maintain stable classification accuracy with an increasing proportion of missing data, as it is crucial to support decision-making. 
This assessment will involve comparing its performance with the previously employed classification with ROC curve abstractions, as well as benchmarking it against established classification algorithms like Gradient Boosting and Random Forest. 
This comparative analysis will determine if the superior PPA abstraction method offers improved accuracy compared to selected classic explainable ML methods and retains robustness in the face of missing data, ultimately elucidating its overall effectiveness within the PPA framework.

\begin{figure}[H]
    \centering
    \includegraphics[width=0.95\linewidth]{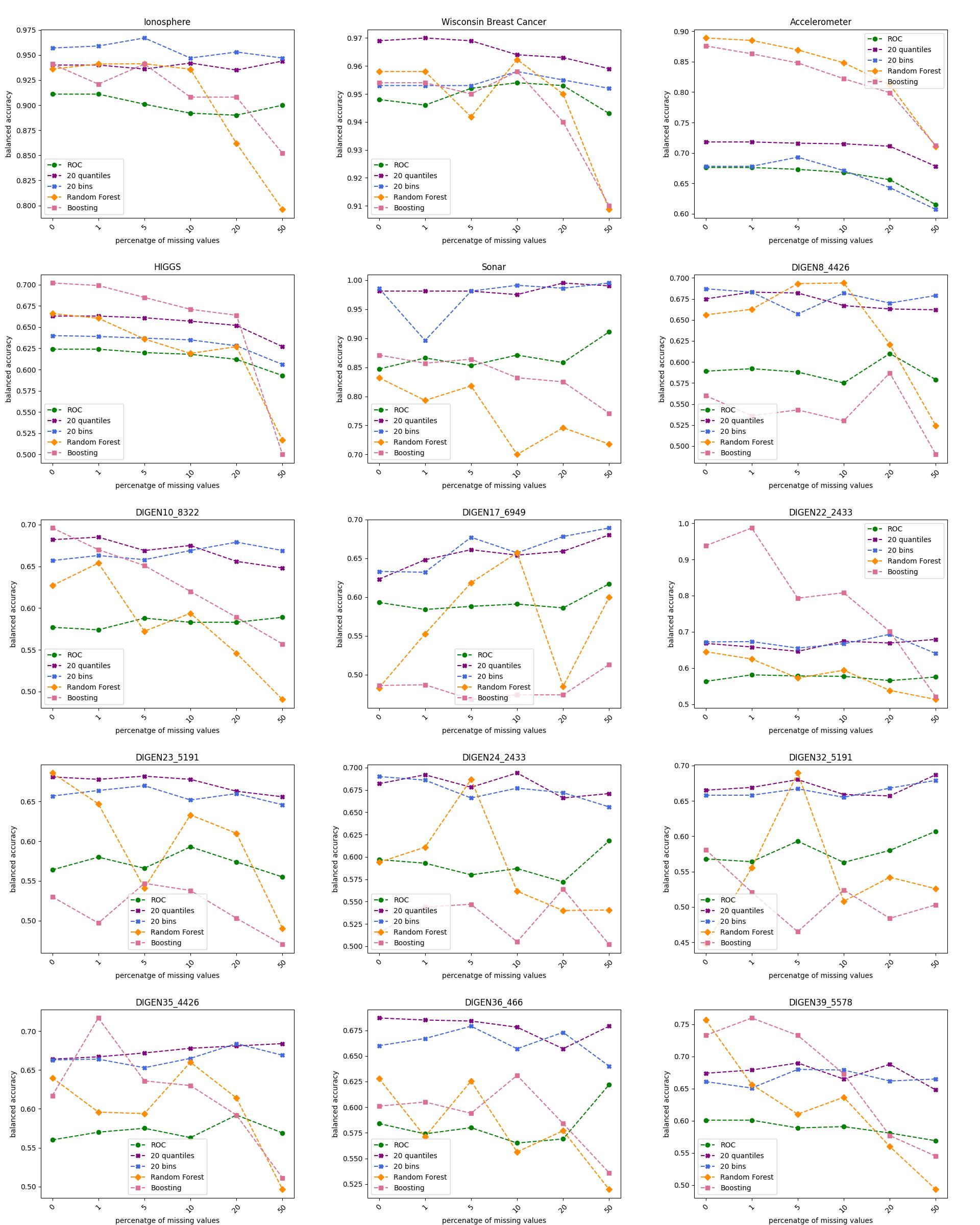}
    \caption{Balanced accuracy (BA) as a function of percentage of missing data for PPA with two best performing abstraction methods (order quantiles 20 and 20 bins), PPA with ROC curve, Random Forest and Gradient Boosting.}
    \label{fig:BA_missing}
\end{figure}

Fig.~\ref{fig:BA_missing} shows the balanced accuracy (see Eq.\eqref{eq:BA}) for PPA with three different abstraction protocols, as well as two state-of-the-art explainable ML methods: RF and GB. 
None of the three presented methods proved to be universally superior across all tested datasets.
However, for complete data, PPA with fixed 20 bins or 20 quantiles achieves the highest BA for 10 datasets.
For $50\%$ of missing data, this number increases to 14, for which only the Accelerometer is not described better using PPA than other methods.
This behaviour can be attributed to the limited number of features, specifically $4$, characterising this dataset. 
Additionally, the size of the dataset, which can be considered average with approximately $10^5$ inputs, favours the efficiency of Random Forest and Gradient Boosting.

Moreover, PPA demonstrated remarkable consistency in accuracy across all abstraction methods and datasets, irrespective of the proportion of missing values. 
In contrast, RF and GB showed a notable decline in performance as the fraction of missing values increased. 
This indicates that PPA offers superior robustness to data quality issues, making it a more reliable choice for analyses involving potentially missing data, which is common in decision-making processes.

BA is not the only important metric from the point of view of classification models. 
Due to its ability to anonymise data and deal with missing values and applicability for small datasets, the proposed ML may be particularly relevant, among other areas, in medicine, social sciences, and finance, where the high level of precision and recall is also very important. 
Therefore, we decided to check whether missing data affects the mentioned metrics for examined models.

\begin{figure}[H]
    \centering
    \includegraphics[width=0.85\linewidth]{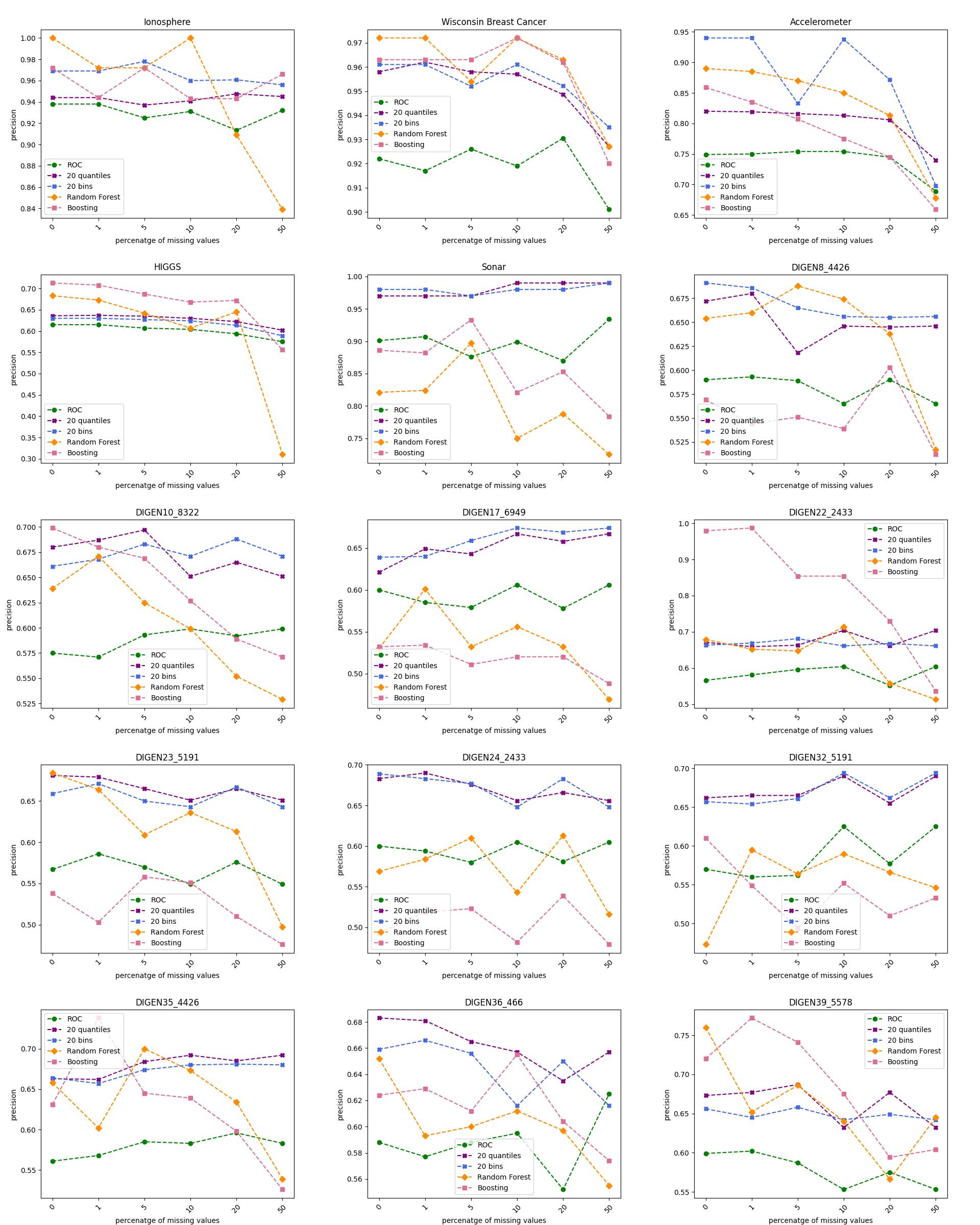}
    \caption{Precision as a function of percentage of missing data for PPA with two best-performing abstraction methods (order quantiles 20 and 20 bins), PPA with ROC curve, Random Forest and Gradient Boosting.}
    \label{fig:Precission_missing}
\end{figure}

\begin{figure}[H]
    \centering
    \includegraphics[width=0.85\linewidth]{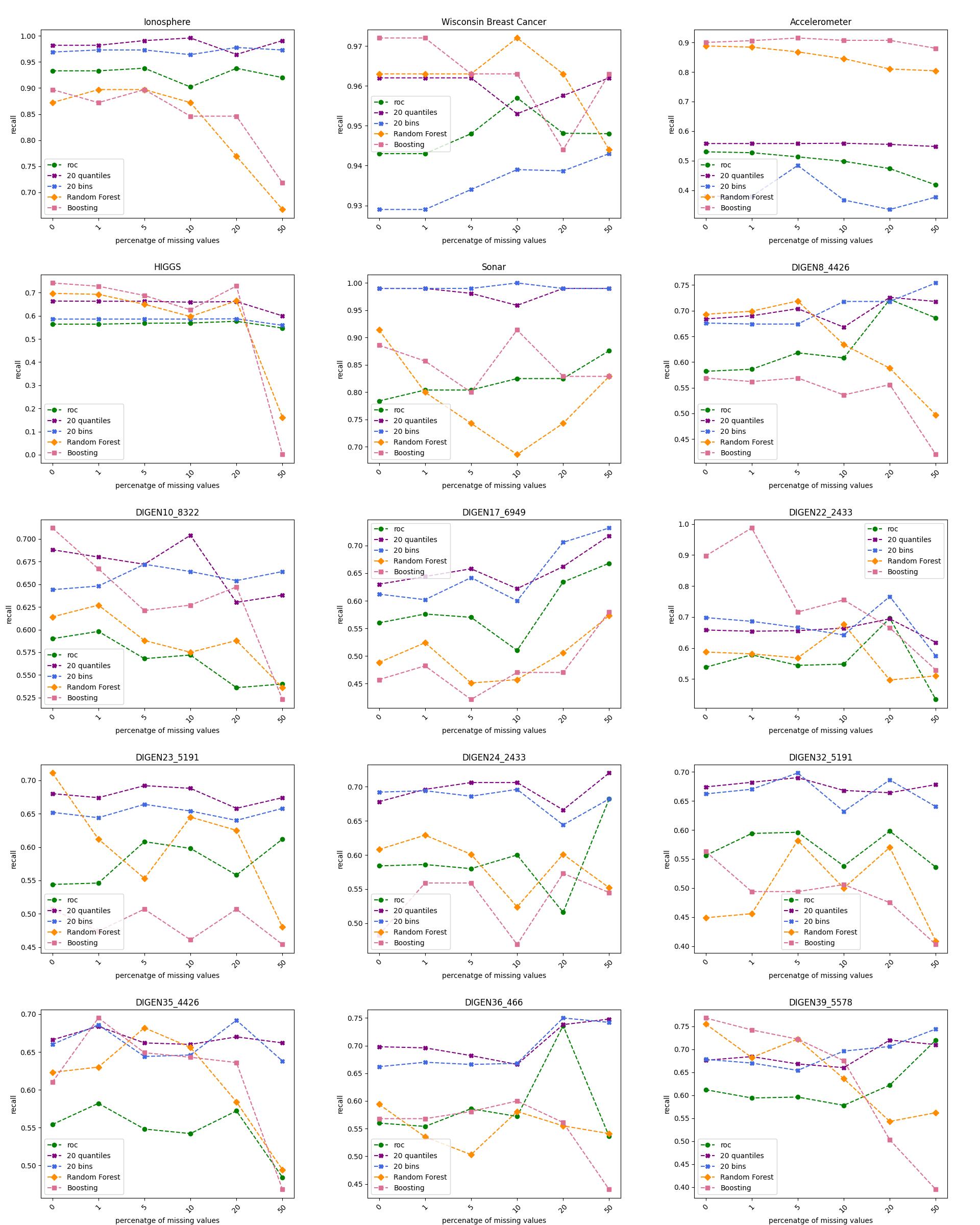}
    \caption{Recall as a function of percentage of missing data for PPA with two best-performing abstraction methods (order quantiles 20 and 20 bins), PPA with ROC curve, Random Forest and Gradient Boosting.}
    \label{fig:recall_missing}
\end{figure}

As presented in Fig.~\ref{fig:Precission_missing}, in the absence of missing data, Random Forest or Gradient Boosting achieves the highest precision values for 7 out of 15 datasets (Ionosphere, Wisconsin Breast Cancer, HIGGS, and DIGENs: 10\_8322, 22\_2433, 23\_5191, 39\_5578), however for DIGEN 23\_5191, the difference was marginal. 
For the rest of the datasets, classification using either quantiles $20$ or $20$ bins yields higher precision. In the case of recall (Fig.~\ref{fig:recall_missing}) Random Forest or Gradient Boosting have an advantage for $8$ datasets (Wisconsin Breast Cancer, Accelerometer, HIGGS and DIGENs: 8\_4426, 10\_8322, 22\_2433, 23\_5191, 39\_5578). 
This suggests that, based on these metrics for the complete datasets, no definitive leader emerges.
With $20$\% of the missing data quantiles $20$ or $20$ bins outperform Random Forest and Gradient Boosting in $12$ out of $15$ cases for precision (Ionosphere, Accelerometer, Sonar and DIGENs: 8\_442, 10\_8322, 17\_6949, 23\_5191, 24\_2433, 32\_5191, 39\_5578, 36\_466, 35\_4426) and in $12$ cases for recall (Ionosphere, Sonar and DIGENs: 8\_442, 10\_8322, 17\_6949, 22\_2433, 23\_5191, 24\_2433,  32\_5191, 35\_4426, 36\_466,39\_5578). The difference was clearly visible for 9 and 10 datasets, respectively for precision and recall.
This indicates the growing advantage of PPA with new abstraction protocols over selected classical models with increasing data uncertainty. 

A greater proportion of missing data works against Random Forest and Gradient Boosting, which is particularly noticeable when $50\%$ of data is removed - in such conditions Random Forest or Gradient Boosting prevails for only 1 dataset; marginal prevalence of Random Forest for precision - DIGEN39\_5578, while clearly visible for both Random Forest and Gradient Boosting for recall - Accelerometer, what was previously noticed for balanced accuracy.

In this section, the performance of three machine learning methods -- RF, GBT, and  PPA -- was compared under increasing levels of missing data. 
The enhanced PPA (ePPA), with new quantile 20 and decile 20 abstractions, demonstrated that BA remains stable even with a high amount of missing data.
This indicates that ePPA achieves better results than the traditional methods in conditions with significant missing data.

Furthermore, recall and precision for these three methods were evaluated as the percentage of missing data increased. 
These metrics were also found to be more stable for ePPA than for the classical methods. 
Given the context of ML supporting decision-making, it is crucial that high and stable values for these three measures are maintained by ePPA even when a substantial amount of data is missing.

\subsection{Feature significance stability}
The final part of the analysis focuses on the stability of significance of the features used for classification. 
It is highly desirable for CDM that the feature's importance for the classification does not change much when missing values are random, i.e., there is no pattern in their appearance.
This can be analysed directly by inspecting values of every feature for every model, however for broad analysis of many models it is convenient to use more comparison-based measures, see Eq.~\eqref{eq:relative_significance} and their moments. 

The analysis of the $50\%$ missing data condition within the stability of the feature selection section was not presented.
This exclusion stems from the observation that all evaluated feature selection methods exhibited substantial instability at this extreme level of incompleteness.  
Including these findings, characterised by high variance, would have hindered the interpretation of the broader trends in stability observed with increasing missing data percentages. 

\begin{figure}[H]
    \centering
    \includegraphics[width=0.75\linewidth]{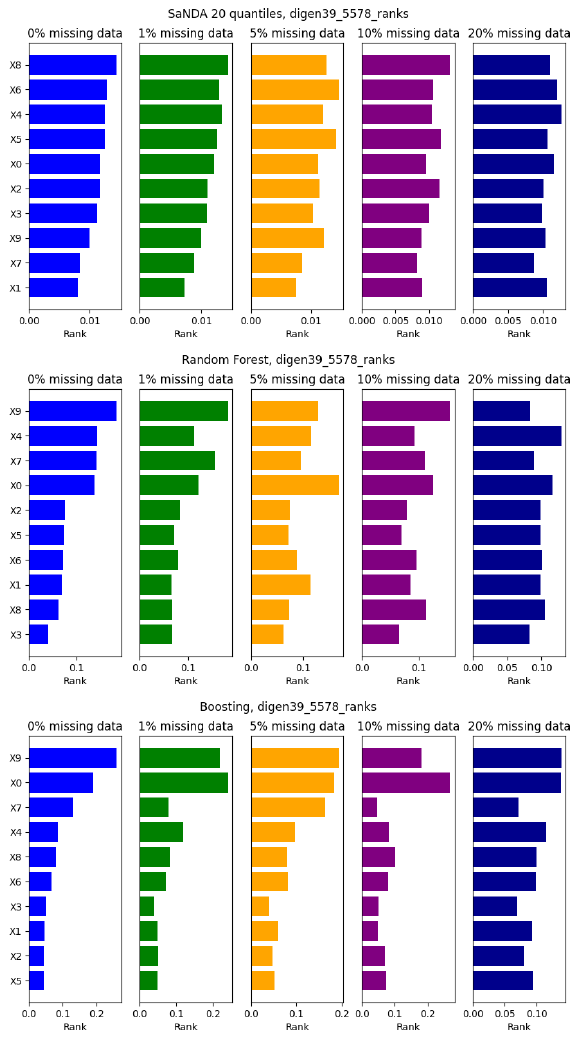}
    \caption{Most important features for the classification of DIGEN39\_5578 for PPA with quantiles 20 abstractions (top), Random Forest (middle) and Gradient Boosting (bottom) for different levels of missing data.}
    \label{fig:featueres_example}
\end{figure}

Fig.~\ref{fig:featueres_example} shows the most important features for the classification of DIGEN39\_5578 depending on the percentage number of missing values. It illustrates how the significance of features in decision-making changes and illustrates the impact of 
incomplete information on the explanatory power of classification. 
The individual bars correspond to the feature importance values for the individual PPA (with deciles of 20), Random Forest, and Gradient Boosting models, from top to bottom, respectively.
They are intended to reflect how important a given feature is for the classification process, and the method of obtaining them is described in the section~\ref{sec:RF} for the considered methods.
Although this approach is inefficient for comparing a large number of data sets, some conclusions can still be drawn, and the results obtained from the analysis of average values can be better understood.
In this dataset, variability is high across all methods. For PPA (top panel), the change in feature importance between $0\%$ and $1\%$ missing data is minimal and requires verification of the values. 
For higher percentages of missing data, changes are noticeable but not critically high. 
In the case of the Random Forest (middle panel), significant changes in feature importance are immediately apparent with the introduction of missing values. 
Additionally, for more than $1\%$ missing data, some features experience changes by more than one-third. 
The behaviour for Gradient Boosting (bottom panel) is similar to that of the Random Forest; however, the less important features remain more stable.
A full representation of feature importance for all analyzed models, datasets and data incompleteness levels is presented in a more compressed form in the supplementary materials.

\begin{figure}[H]
    \centering
    \includegraphics[width=0.95\linewidth]{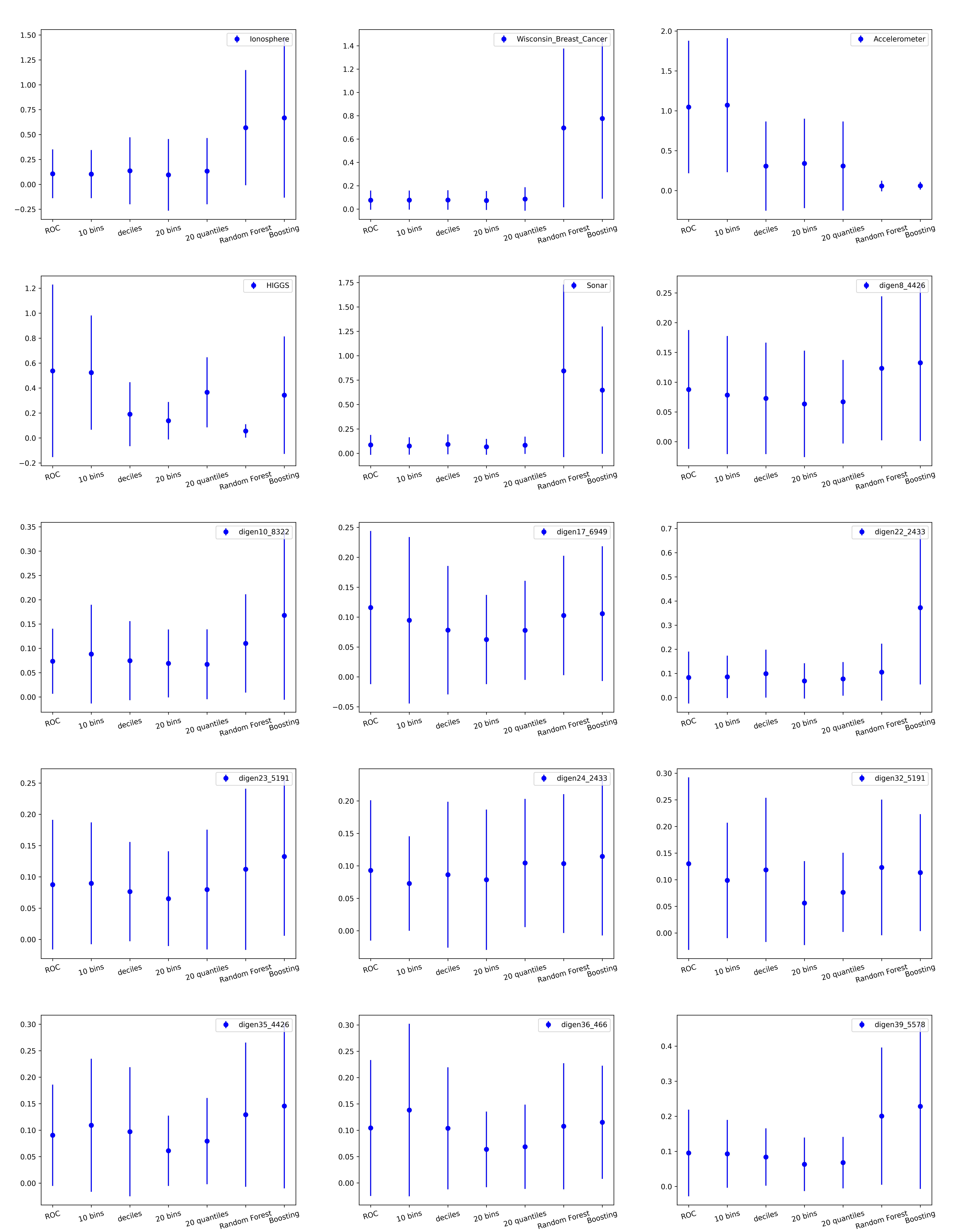}
    \caption{The average and standard deviation of the relative feature significance change given by Eq.~\eqref{eq:relative_significance} for various amount of missing data.}
    \label{fig:Averaged_difference_between_features}
\end{figure}

Fig.~\ref{fig:Averaged_difference_between_features} shows the average relative feature importance change for each dataset when using different ML models. 
From the construction of the measured quantity, a small value indicates that the features do not undergo significant changes. 
Thus, a low mean suggests that all features change minimally when transitioning from complete to incomplete data. However, the mean alone provides limited insight into the behaviour. 
Therefore, the points corresponding to the average are supplemented with error bars representing the standard deviation of the measure used. 
These error bars provide insight into the spread of the change in feature importance. 
A low standard deviation, in addition to a low mean, indicates a model where the significance of features is robust to missing data. 
This is because a low standard deviation means that the majority of features do not deviate significantly from the mean, indicating that it is not a case where some features change significantly while others do not.

Based on the above quantities, three scenarios can be distinguished, corresponding to the increasing level of feature importance stability. 
First, a high mean with high variance indicates that changes are virtually random, and potential matches are a matter of chance. 
Second, a state of low mean change but high variance suggests that relative consistency can be expected in the importance of features, although clearly visible changes in feature importance may still appear even with less than $20\%$ of missing data. 
Finally, when a low mean is combined with a low standard deviation, it indicates cases where the model does not change much even with $20\%$ of missing values, so it represents the stability of the important features across a variety of percentage numbers of missing data.

All three of the above scenarios can be observed for the models shown in the Fig.~\ref{fig:Averaged_difference_between_features}.
For two DIGEN datasets, all employed methods demonstrate high mean values and high standard deviations.
For the remaining DIGEN datasets, at least one abstraction method used in PPA shows a lower mean than Gradient Boosting or Random Forest; however, the standard deviation generally remains large. 
Notably, for three DIGEN datasets, as well as Sonar, WDBC, and Ionosphere, PPA consistently exhibits a lower mean and a smaller standard deviation than the other methods. 
In the Higgs dataset, Random Forest achieves a low mean and standard deviation, while the other methods perform worse, although some PPA abstractions display only slightly higher mean and standard deviation. 
For the accelerometer dataset, traditional methods achieve low mean and variance, while PPA performs poorly. 
This underperformance of PPA is attributed to previously discussed issues: the low number of features is unfavourable for PPA, while the large volume of data benefits the traditional ML methods.
Overall, these results highlight the stability and reliability of the most important features identified by PPA, particularly in datasets like Sonar, WDBC, and Ionosphere, where it consistently outperforms traditional methods.

\section{Summary and Conclusions}\label{sec:summary}
Incomplete data, a prevalent challenge for decision-making in critical domains, significantly hinders tasks like classification and feature importance analysis. 
Traditional machine learning (ML) approaches effectively handle small amounts of missing data through imputation or modelling techniques~\cite{awadeh2022evoimputer}. 
However, their performance significantly degrades when dealing with datasets exhibiting high levels of incompleteness. 
To address this challenge, we recently proposed algorithm (PPA)~\cite{ibias2023sanda,gherardini2024cactus} which demonstrates remarkable stability in classification performance even with increasing data incompleteness. 
This characteristic makes PPA a promising candidate for critical decision-making tasks.  
A limitation of PPA, however, is its lower accuracy, particularly when applied to complete datasets, where its performance falls below that of other explainable AI methods.  The first part of the work focuses on enhancing PPA's performance by modifying its method for abstraction generation. 
Subsequently, a comparative analysis is conducted to evaluate the utility of three methods – PPA, Random Forest (RF), and Gradient Boosting (GB) – under varying degrees of missing data.

The impact of modifying PPA's abstraction generation method on its effectiveness was investigated experimentally. 
This modification involved segmenting the data space into partitions of fixed width and cardinality (number of elements). 
Abstractions with fixed cardinality generally exhibited superior performance compared to alternative approaches. 
However, a more substantial improvement in model accuracy was observed with an increase in the total number of abstractions employed.

The selection of the number of abstractions directly connects to the trade-off between achieving high accuracy and effectively managing uncertainty. 
Determining the optimal number of abstractions necessitates careful consideration of the specific problem domain. 
This includes defining acceptable error tolerances and the desired level of precision.

An examination of PPA with modified abstractions reveals balanced accuracy, precision, and recall comparable to popular critical decision-making ML algorithms, such as Gradient Boosting and Random Forest, even with complete data. 
Notably, PPA's accuracy degrades significantly slower with data incompleteness, remaining high even with $50\%$ missing data. 
These findings suggest PPA with modified abstractions as a viable alternative for critical decision-making tasks, particularly due to its robustness to missing data. 

Missing data, particularly in the most important features, poses a significant challenge for classification tasks by introducing variability into the feature importance.
Even with a random distribution of missing values, their presence can significantly alter the relative importance of individual features for the final classification outcome. 
This is undesirable, as the model should ideally reflect the underlying population characteristics.
Therefore, a desirable property for the classification model is low sensitivity to changes that do not substantially disrupt the data structure or introduce bias into the dataset.

A comparison of feature importance changes across PPA, RF, and GB reveals a significant advantage for PPA. The relative importance of individual features for classification remains largely stable with PPA even for small to moderate levels of missing data. Conversely, the most important features for classification using Gradient Boosting and Random Forest exhibit drastic changes in importance with even minimal ($1\%$) missing data. This suggests that PPA exhibits superior robustness to missing data in terms of feature importance for classification tasks.  Similar resilience was also observed for changes in feature order.

\section*{Acknowledgements}
This research has been supported by the European Union’s Horizon $2020$ research and innovation programme under grant agreement Sano No. $857533$ on the basis of the contract No. MEiN/2023/DIR/3796.
This publication is supported by Sano project carried out within the International Research Agendas programme of the Foundation for Polish Science MAB PLUS/2019/13, co-financed by the European Union under the European Regional Development Fund.
This research was supported in part by PLGrid Infrastructure.

Suggestions from Jan Argasi\'nski are greatly acknowledged.

\end{document}